\def\year{2021}\relax
\documentclass[letterpaper]{article} % DO NOT CHANGE THIS
\usepackage{aaai21}  % DO NOT CHANGE THIS
\usepackage{times}  % DO NOT CHANGE THIS
\usepackage{helvet} % DO NOT CHANGE THIS
\usepackage{courier}  % DO NOT CHANGE THIS
\usepackage[hyphens]{url}  % DO NOT CHANGE THIS
\usepackage{graphicx} % DO NOT CHANGE THIS
\usepackage{graphicx}  % DO NOT CHANGE THIS
\usepackage{natbib}  % DO NOT CHANGE THIS AND DO NOT ADD ANY OPTIONS TO IT
\usepackage{caption2} % DO NOT CHANGE THIS AND DO NOT ADD ANY OPTIONS TO IT
\nocopyright
\title{Disentangled Motif-aware Graph Learning for Phrase Grounding}
\author{
    Zongshen Mu\textsuperscript{\rm 1},
    Siliang Tang\textsuperscript{\rm 1}\thanks{corresponding author},
    Jie Tan\textsuperscript{\rm 1},
    Qiang Yu\textsuperscript{\rm 2},
    Yueting Zhuang\textsuperscript{\rm 1}
    \\
}
\affiliations{
    \textsuperscript{\rm 1}DCD Lab, College of Computer Science, Zhejiang University\\
    \textsuperscript{\rm 2}City Cloud Technology (China) Co., Ltd\\
    \{zongshen,siliang,tanjie95,yzhuang\}@zju.edu.cn\qquad
    yq@citycloud.com.cn
}

\usepackage{amsmath}
\usepackage{amssymb}
\usepackage{algorithm}
\usepackage{algorithmic}
\usepackage{booktabs}
\usepackage{enumitem}
\usepackage{multirow}
\usepackage[switch]{lineno}

\begin{document}
%\linenumbers

\nocopyright

\maketitle

\begin{abstract}
In this paper, we propose a novel graph learning framework for phrase grounding in the image. Developing from the sequential to the dense graph model, existing works capture coarse-grained context but fail to distinguish the diversity of context among phrases and image regions. In contrast, we pay special attention to different motifs implied in the context of the scene graph and devise the disentangled graph network to integrate the motif-aware contextual information into representations. Besides, we adopt interventional strategies at the feature and the structure levels to consolidate and generalize representations. Finally, the cross-modal attention network is utilized to fuse intra-modal features, where each phrase can be computed similarity with regions to select the best-grounded one. We validate the efficiency of disentangled and interventional graph network (DIGN) through a series of ablation studies, and our model achieves state-of-the-art performance on Flickr30K Entities and ReferIt Game benchmarks.

\end{abstract}

\section{Introduction}
\begin{figure}[t]
	\centering
	\includegraphics[width=1\columnwidth]{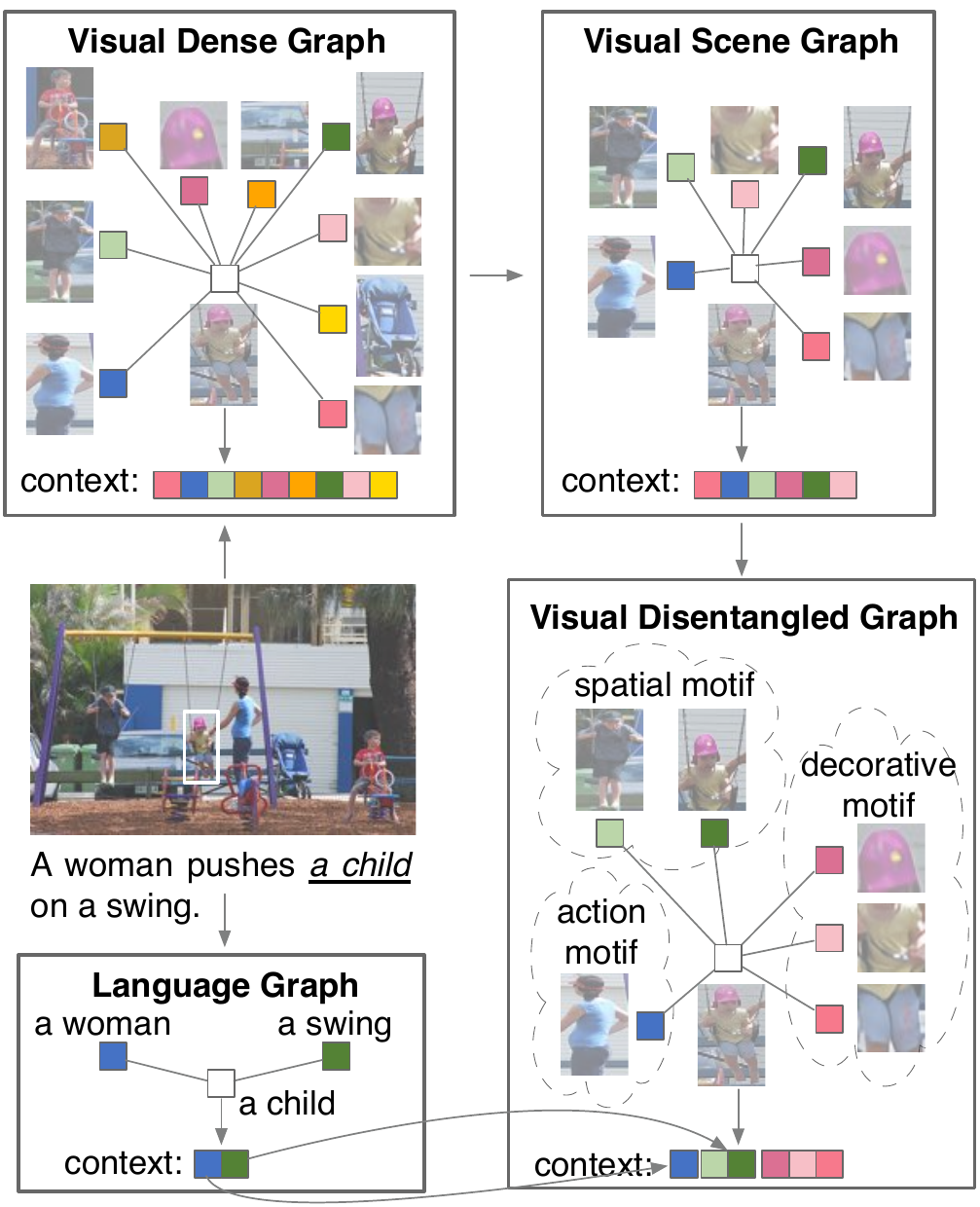}
	\renewcommand{\captionlabeldelim}{. }
	\vspace{-0.3cm}
	\caption{An example of phrase grounding.}
	\vspace{-0.3cm}
	\label{fig:dis_example}
\end{figure}

\begin{figure*}[ht]
	\includegraphics[width=1\textwidth]{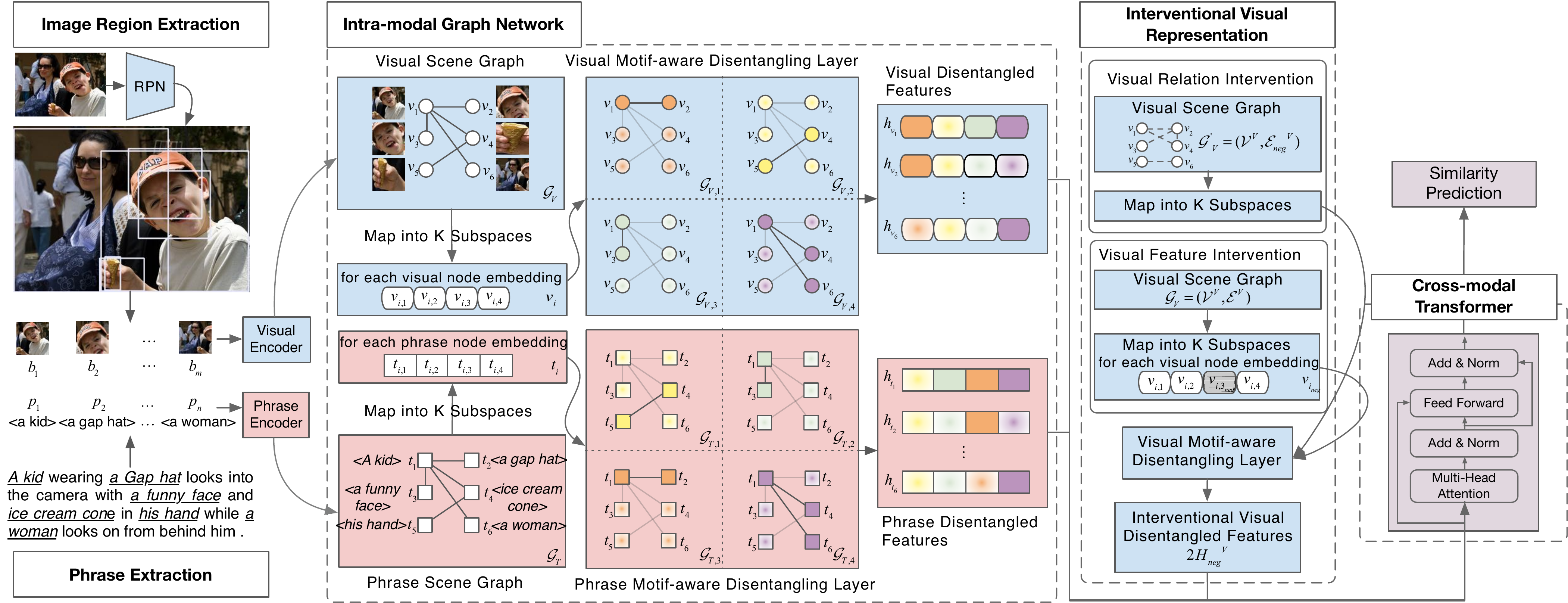}
	\renewcommand{\captionlabeldelim}{. }
	\vspace{-0.3cm}
	\caption{DIGN architecture. Node colors interpret different motifs' context, and shades of color reveal the motif's importance.}
	\vspace{-0.3cm}
	\label{fig1}
\end{figure*}

On giving an image-sentence pair, the Phrase Grounding (or more generally, Language Grounding) task aims to ground multiple noun phrases from the given sentence to the corresponding image regions, as illustrated in Figure \ref{fig1}. By aligning textual modality to visual modality, phrase grounding is able to link knowledge from both modalities and improves multimodal tasks that span visual question answering, visual commonsense reasoning, and robotic navigation.

Since the information that the image usually contains is more than that of the sentence, the general solution is to understand the surrounding visual context of a visual object, especially pay attention to the context considered by the sentence as well, and then map it to its referent words.
Nevertheless, previous studies fail to consider such fine-grained context. Some methods neglect the importance of visual or textual context, nor do they ground individual phrases independently \cite{plummer2018conditional} or sequentially \cite{dogan2019neural, li2020does}. This leads to poor performance when objects with completely different functions or actions have the similar visual effect in the same image. More recently, \citet{bajaj2019g3raphground} and \citet{liu2020learning} utilize fully connected dense graphs to capture coarse-grained context among visual objects, and improve grounding results. 

However, their graphs have three fatal problems. First, the dense graph holds noisy false connections and makes errors in complex scenes. Second, they treat all the relations equally. Under such a framework, the node (i.e., phrase or visual object) representation is a mixture of all the possible relations. In reality, it is insufficient to reveal various high-order relation categories between nodes, or motifs\footnote[1]{ We post-hoc motifs referring to the method in the topic model.} (e.g., \emph{decorative}, \emph{part-of}, \emph{spatial}, \emph{action}), in the contextual information.  For instance, as shown in Figure \ref{fig:dis_example}, ``a child" needs to be grounded in the image that includes multiple similar object regions. As mentioned above, the language graph (the lower left of Fig. \ref{fig:dis_example}) is only the subgraph of visual contents. When we learn the representation of ``a child" in the dense graph (the upper left of Fig. \ref{fig:dis_example}), it is difficult to ground the right region in noisy context. After the spurious neighbors are filtered in the scene graph (the upper right of Fig. \ref{fig:dis_example}), it is also difficult to make a grounding decision because too many redundant motifs around the child prevent the model from grasping the vital motif. Third, there are some co-occurring visual objects in the datasets, while some of irrelevant co-occurrences are of high probability \cite{wang2020visual}, which influences the graph structure and finally damages the model's robustness and generalization capability.

In this paper, we propose the Disentangled Interventional Graph Network (DIGN),  a more explainable framework for phrase grounding. First, instead of using a fully connected dense graph, we use two scene graphs to model the context among phrases and regions. Then, we use two disentangled graph networks to integrate these various context into disentangled dimensions of node (i.e., phrase or visual object) representation. Each chunk, referred to as a motif-aware embedding, can represent the strength of a certain motif (the lower right of Fig. \ref{fig:dis_example}), which greatly reduces the noise from spurious connections and creates fine-grained intra-modal representations for later inter-modal fusion. Furthermore, we propose the interventional visual sample synthesizing training scheme to alleviate the co-occurring biases problem and improve the model's robustness.

Our key contributions are threefold:
\begin{itemize}
	\item We devise motif-aware graph learning based on disentangled graph network, which distinguishes diverse context and considers fine-grained contextual fusing.
	\item We adopt the feature and the structure interventional strategies to learn robust representations in the graph.
	\item We conduct a series of experiments that demonstrate our DIGN model achieves more competitive performance than the prior state-of-the-art.
\end{itemize}
 
\section{Related Work}
There are two lines of literature that are closely related to our work: phrase grounding and disentangling representation.

\subsection{Phrase Grounidng}\label{phrase_related}
To ground the spatial region described by the phrase in an image, early methods \cite{wang2016learning, wang2018learning, plummer2018conditional, akbari2019multi, plummer2020shapeshifter} learn a shared embedding subspace and compute the distance of each phrase-region pair. The aforementioned  methods ground each phrase independently and neglect textual and visual context information. One step forward, the RNN structure \cite{hu2016natural, dogan2019neural} and the Transformer model \cite{li2020does} are used to capture context among phrases and regions. To formulate more complex and non-sequential dependencies, graph architecture \cite{bajaj2019g3raphground, liu2020learning} becomes increasingly prevalent and leverage relationships between nodes to enrich representations and achieve alignment. They treat the context uniformly at a holistic aspect, which results in suboptimal representation and limited interpretability. An underlying fact is omitted that: grounding the phrase to the region is affected by different motifs in the graph's context. Our model disentangles context at a fine-grained level and incorporates motif-aware context into the representation.

\subsection{Disentangling Representation}
Learning to identify and disentangle explanatory factors hidden in the observed milieu of sensory data is the foundation of artificial intelligence. Early attempts \cite{tishby2015deep} apply the information bottleneck method only to capture and extract a relevant or meaningful data summary. Based on information-theoretic concepts, $\beta$-VAE \cite{Higgins2017betaVAELB} can learn disentangled representations of independent visual data generative factors in a completely unsupervised manner. Later studies \cite{burgess2018understanding, chen2018isolating} achieve further improvements for disentangled factor learning at a more granular level. These existing efforts are made to discover different latent factors on non-relational data. More recently, disentangled representation learning is explored in the filed of graph-structured data \cite{ma2019disentangled, zheng2020disentangling}, most of which mainly disentangle latent factors behind user behavior in bipartite graphs commonly seen in recommender systems. This work focuses on not only disentangling motifs in the intra-modal scene graph but also keeping disentangled in cross-modal common subspace. The disentangled graph network for phrase grounding is related to previous modular method, MattNet \cite{yu2018mattnet}, but it is implicit to assign the meaning of the chunked representation and convenient to transfer to other tasks.

\section{Approach}
Given multiple phrases and the corresponding image, we denote a set of noun phrases as $\mathcal{P}=\{p_i\}_{i=1}^n$ and their spatial regions (or bounding boxes) as $\mathcal{B}^{\text{+}}=\{b^{\text{+}}_i\}_{i=1}^n$ where $b^{\text{+}}_i=(x_1,y_1,x_2,y_2)$ is the left-top and right-bottom coordinates. The primary goal is to select N bounding boxes from region proposals $\mathcal{B}=\{b_i\}_{i=1}^m(m>>n)$ as much overlap with ground truth annotations $\mathcal{B}^{\text{+}}$ as possible and get grounding results.

 Figure \ref{fig1} illustrates our disentangled interventional graph network (DIGN). The framework are composed of four stages to implement phrase grounding at a fine-grained level: 1) phrase and visual feature encoding; 2) two intra-modal disentangled graph networks; 3) interventional operators in the visual graph; 4) a cross-modal Transformer. 
 
\subsection{Phrase and Visual Encoders}
Each phrase consists of a word or a sequence of words. For phrase features, we encode each word of the phrase to a real-valued vector by a pre-trained BERT \cite{Devlin2019BERTPO}, then compute the phrase representation by taking the sum of word embeddings in each $p_i$. In respect of visual features, we use a region proposal network \cite{ren2015faster} to extract the top-M proposals with diversity and discrimination. Each proposal $b_i$ is represented after the last fully-connected layer of RoI-Align pooling \cite{he2017mask}. To transform phrase and visual features to the identical dimensions, $t_i\in\mathbb{R}^{d^t_{\text{in}}}$ and $v_i\in\mathbb{R}^{d^v_{\text{in}}}$ are obtained severally through two fully-connected layers combined with the ReLU activation function and a batch normalization layer.

\subsection{Intra-Modal Graph Network}
To separate various context among noun phrases and region proposals and provide explicit channels to guide motif-aware information flow, we introduce the following two intra-modal disentangled graph networks below.

\subsubsection{Phrase Disentangled Graph}
Instead of a noisy dense graph to capture linguistic context, we construct a phrase scene graph $\mathcal{G}_T=(\mathcal{V}^T,\mathcal{E}^T)$ from the image caption including phrases by an off-the-shelf scene graph parser \cite{schuster2015generating}. Formally, $\mathcal{V}^T=\{t_i\}_{i=1}^n$ represents nodes equipped with phrase embeddings, $(i, j)\in \mathcal{E}^T$ indicates that there is an edge connecting $t_i$ and $t_j$, and $\mathcal{N}^T_i=\{t_j\colon(i,j)\in\mathcal{E}^T\}$ denotes the neighbors of $t_i$. 

When grounding the phrase to the image region, the situation that similar object regions are required to be differentiated is universal and intractable. Intuitively, one phrase interacts with neighbors and makes use of its context to distinguish the regions. However, different situations need to consider the different context of the neighborhood. As shown in Figure \ref{fig1}, $t_1$ links with $t_2$, $t_3$, $t_4$, and $t_6$, whose context includes ``wearing", ``with", ``in", and ``look on from behind" high-order relation patterns. The context can be categorized into \emph{decorative}, \emph{part-of}, \emph{spatial}, and \emph{action} motifs respectively. Clearly, both the kid and the woman have funny faces, but a funny face is grounded to the kid because of his unique decorative motif (e.g., he has a gap hat) and spatial motif (e.g., the ice cream cone is on his hand), which do not belong to the context of the woman.

Therefore, unlike standard graph convolutional network \cite{velivckovic2017graph, DBLP:conf/iclr/XuHLJ19} viewing context of the ego node as a holistic one to message and update, we aim to learn a disentangled representation over various motifs, as follows:
	\begin{equation}
		h^T_i=[h^T_{i,1},h^T_{i,2},...,h^T_{i,K}]\label{eq1}
	\end{equation}
	where $h^T_i$ is the final output of $t_i$, $K$ is the hyperparameter controlling the number of latent motifs. It is expected that $h^T_{i,k}\in \mathbb{R}^{\frac{d^T_{\text{out}}}{K}}$ incorporates the contextual information of the $k\text{-th}$ motif and is independent of other components. For the $k\text{-th}$ chunked embeddings of all nodes, we define a motif-aware graph as $\mathcal{G}_{T,k}=\{\mathcal{V}^T_k,\mathcal{E}^T\}$. As such, we build a set of subgraphs $\mathcal{G}_T=\{\mathcal{G}_{T,1},\mathcal{G}_{T,2},...,\mathcal{G}_{T,K}\}$ to refine motif-aware linguistic features through graph disentangling layers.

To separate the original mixed embedding, we project $t_i$ into $K$ subspaces, as shown below:
	\begin{equation}
		%t^0_{i, k}=\sigma({W^T_k}^0 t_i+{b^T_k}^0)/\left\|\sigma({W^T_k}^0 t_i+{b^T_k}^0)\right\|_2
		t^0_{i, k}=\frac{\sigma({W^T_k}^0 t_i+{b^T_k}^0)}{\left\|\sigma({W^T_k}^0 t_i+{b^T_k}^0)\right\|_2}\\
	\end{equation}
	where ${W^T_k}^0\in \mathbb{R}^{\frac{d^T_{\text{out}}}{K} \times d^T_{\text{in}}}$ and ${b^T_k}^0 \in \mathbb{R}^{\frac{d^T_{\text{out}}}{K}}$ are the parameters of channel $k$, the superscript 0 shows the original disentangled state; and $\sigma(\cdot)$ is a nonlinear activation function. We use $l_2$-normalization to ensure numerical stability and prevent the neighbors with overly rich features from distorting our prediction. Other components of $t_i$ is computed as $t^0_{i, k}$, which are fed to the corresponding subgraph $\mathcal{G}_{T,k}$.

After initializing nodes of motif-aware subgraphs, we distill useful information of neighbors to comprehensively capture $K$ latent factors' context. For each motif-aware subgraph, we design a phrase graph disentangling layer:
	\begin{align}
		t^1_{i, k}&=\sigma({W^T_{ek}}^1 t^0_{i, k}+\sum\nolimits_{j\in \mathcal{N}_i^T}{a^T_{j, k}}^1{W^T_k}^1 t^0_{j, k})\\
		%{a^T_{j, k}}^1&=exp({t^0_{j, k}}^\intercal t^0_{i, k})/\sum\nolimits_{k'=1}^K exp({t^0_{j, k'}}^\intercal t^0_{i, k'}) \label{eq_routing}
		{a^T_{j, k}}^1&=\frac{\text{exp}({t^0_{j, k}}^\intercal t^0_{i, k})}{\sum\limits_{k'=1}^K \text{exp}({t^0_{j, k'}}^\intercal t^0_{i, k'})} \label{eq_routing}
	\end{align}
	where ${a^T_{j, k}}^1$ $({a^T_{j, k}}^1\geq0, \sum\nolimits_{k'=1}^K {a^T_{j, k}}^1=1)$ provide the $k\text{-th}$ motif's importance why $t_j$ connects with $t_i$; $\mathcal{N}_i$ is the first-order neighbors of $t_i$; ${W^T_{ek}}^1$ and ${W^T_k}^1 \in \mathbb{R}^{\frac{d^T_{\text{out}}}{K} \times \frac{d^T_{\text{out}}}{K}}$ represent the learnable parameters of the ego node and neighbor nodes at the $k\text{-th}$ aspect respectively; the superscript 1 represents the output of the graph disentangling layer from one-hop neighbors; and $\sigma$ is ReLU activation function followed by dropout with probability 0.5. 

Equation \ref{eq_routing} is the key step to the first-order neighbor routing. When calculating it with all neighbors, we can get the contextual information for motif $k$, which is similar to the attention mechanism, but not exactly the same. The attention mechanism like Graph Attentive networks \cite{wang2019neighbourhood} coarsely aggregates one-hop neighbors and updates the ego node as an entirety, causing difficult to distinguish different context and entangling latent motifs. Conversely, the neighbor rooting mechanism slices the node into $k$ components, for which each part can attend similar context and provide an interpretable result.

Having used one-hop neighbors, we can stack more graph disentangling layers to gather influential signals from higher-order neighbors. After $L$ layers, we sum up the motif-aware representations at different layer as the $k\text{-th}$ chunk representation:
	\begin{equation}
		h^T_{i, k}=t^0_{i, k}+t^1_{i, k}+...+t^L_{i, k}
	\end{equation}
	We argue that $h^T_{i, k}$ describes the semantic context related with the $k\text{-th}$ motif. Finally, we joint $K$ chunks to update $t_i$ after $L$ disentangling layers, as presented in equation \ref{eq1}. Consequently, we not only disentangle phrase representations but also explain each part of them.

\subsubsection{Visual Disentangled Graph}
We establish a visual scene graph $\mathcal{G}_V=(\mathcal{V}^V,\mathcal{E}^V)$ by the scene graph generation model \cite{zellers2018neural} that exploits visual relationships among region proposals.

We hypothesize that the visual graph context implies the same categories of motifs as those in the context of the phrase graph. So we also slice the visual node embedding into $K$ components, and every chunked node embeddings constructs independent subgraph $\mathcal{G}_{V,k}=(\mathcal{V}^V_{k},\mathcal{E}^V)$. $\mathcal{G}_{V}$ is composed of $K$ motif-aware graphs $\{\mathcal{G}_{V,1},\mathcal{G}_{V,2},...,\mathcal{G}_{V,K}\}$. Similar to the phrase graph network, we use another disentangled graph network to integrate visual context containing varying motifs into representations. More details are presented in Algorithm 1 of Appendix A. 

We firstly map each $v_i$ into $K$ subspaces to initialize $K$ chunked embeddings by linear transformations:
	\begin{equation}
		%v^{0}_{i, k}=\sigma({W^V_k}^0 v_i+{b^V_k}^0)/\left\|\sigma({W^V_k}^0 v_i+{b^V_k}^0)\right\|_2
		v^{0}_{i, k}=\frac{\sigma({W^V_k}^0 v_i+{b^V_k}^0)}{\left\|\sigma({W^V_k}^0 v_i+{b^V_k}^0)\right\|_2}
	\end{equation}
	where ${W^V_k}^0\in \mathbb{R}^{{\frac{d^V_{\text{out}}}{K} \times d^V_{\text{in}}}}$ and ${b^V_k}^0 \in \mathbb{R}^{\frac{d^V_{\text{out}}}{K}}$, $v^0_{i, k}$ is the $k\text{-th}$ part of $v^0_{i}$. The aggregation and updating steps of a visual graph disentangling layer about the $k\text{-th}$ motif can be described as:
	\begin{equation}
		v^1_{i, k}=\sigma({W^V_{ek}}^1 v^0_{i, k}+\sum\nolimits_{j\in \mathcal{N}_i^V}{a^V_{j, k}}^1{W^V_k}^1 v^0_{j, k})
	\end{equation}
	where ${W^V_{ek}}^1$, ${W^V_k}^1 \in \mathbb{R}^{{\frac{d^V_{\text{out}}}{K} \times \frac{d^V_{\text{out}}}{K}}}$, ${a^V_{j, k}}^1$ is the result of the neighbor rooting mechanism in visual graph network. After iterating more layers, the final output of each chunk $h^V_{i, k}$ is obtained by summing up every layers' hidden states of $v_i$:
	\begin{equation}
		h^V_{i, k}=v^0_{i, k}+v^1_{i, k}+...+v^L_{i, k}
	\end{equation}
	The disentangled result $h^V_i$ is depicted to:
	\begin{equation}
		h^V_i=[h^V_{i,1},h^V_{i,2},...,h^V_{i,K}]
	\end{equation}

\subsection{Interventional Visual Representation}
The dataset co-occurring bias, the irrelevant relations, may lead to in-correct attention. When modeling the context of regions in the image by disentangled graph network, it is possible that the biased data plagues chunked representation with irrelevant information through context aggregation. Notably, visual features from the encoder are vulnerable to learn the bias \cite{cadene2019rubi}. Thus, we intervene in representation learning in disentangled graph network to reduce the effect of bias and robust representations.

The core idea of intervention is to change the environment, helping the model train with more unseen data and discover more possible reasons. We utilize the interventional strategies to improve motif-aware learning during the training process. It is efficient and effective to construct interventional samples at the feature and the structure levels in the graph-structure data. Compared to the masked attentive region strategy \cite{liu2019improving}, our strategies do not depend on language attention and are more flexible to learn context-aware features in the graph. The pseudocode of the interventional process is listed in Appendix B.
	
We first introduce the strategy of structure intervention. We aim to change the original edges in the visual scene graph. When modifying the nodes of edges, we just change the target node of the directional edge. In this way, it can eliminate the impact of changing other connected nodes' features as much as possible. We interrupt and randomly interchange all the target nodes' edges in the visual graph. Integrated unrelated motif context into the representation by the neighbor routing mechanism, these contrastive samples help the network to grasp natural motifs between nodes.

The feature interventional samples are synthesized with the implementation of masking right chunked features with incorrect data. There are two types of masked feature interventional methods. One type is to replace features of neighbor nodes in corresponding dims. Neighbors are usually analogous to the ego node on the semantic aspect, making the model distinguish similar cases. The other type is to fill each dim of chunked embedding with noisy distribution. The generated features are treated as noisy training data. It is beneficial and general for the model to learn more unseen samples and have the resilience to noisy data. Therefore, the interventional samples can promote the model to find and understand the meaning of the replaced chunk automatically for phrase grounding.

We define the particular loss function in Section \ref{training} to make full use of interventional samples. Moreover, the intervention module is vital to motif-aware learning, which is demonstrated in the following ablation studies. 

\subsection{Cross-Modal Transformer}
To align two modal representations in a shared space, we introduce a cross-modal Transformer, as shown in the last column of Figure \ref{fig1}. Unlike simple concatenation of two modal features, the multi-head attention mechanism can benefit each head block to capture the corresponding motif's mutual information from both modal disentangled representations. When the visual information is injected to the linguistic encoding and the linguistic information is incorporated to the visual encoding, the Transformer can dynamically select judicious cues for target representation.

We denote the inputs for the Transformer as $h^T=\{h^T_1,h^T_2,...,h^T_n\}$ and $h^V=\{h^V_1,h^V_2,...,h^V_m\}$, which are processed by phrase and visual graph networks respectively. Specifically, multi-head attention is computed as $\text{Multihead}(Q,K,V)$, where $Q$ is the query, $K$ is the key, and $V$ is the value. The details can refer to \citet{vaswani2017attention}. Two modal features are interacted by:
	\begin{align}
		c^T&=\text{Multihead}(W_q h^T,W_k h^V,W_v h^V)\\
		c^V&=\text{Multihead}(W_q h^V,W_k h^T,W_v h^T)
	\end{align}
	where $W_q$, $W_k$ and $W_v$ are learnable weights, $c^T$ is the blended feature from phrases, while $c^V$ is the guided representation from regions.
	
Through this form Transformer, linguistic and visual features are associated so that the similarity prediction is defined as:
	\begin{equation}
		\text{sim}(c^T_i, c^V_j)=c^T_{i}\cdot c^V_{j}
	\end{equation}
	where we just use an inner (dot) product to achieve state-of-the-art performance. The similarity score is thought as the the probability whether  the phrase $p_i$ is grounded to the image region $b_i$.
		
\subsection{Training Objectives}\label{training}
Our loss function includes three parts. The first part is the phrase independent loss. For encouraging $K$ components of the node embedding to be independent, we apply distance correlation \cite{szekely2007measuring} to characterizing independence of any two paired vectors. Each pair's coefficient is zero if and only if these vectors are independent. After phrase disentangled graph network, we formulate the independent loss of phrase chunked embeddings as:
	\begin{align}
		&{\ell}^T_{ind}=\sum\nolimits_{i=1}^n \sum\nolimits_{k=1}^K \sum\nolimits_{k'=k+1}^K \rho (h^T_{i, k}, h^T_{i, k'})\\
		&\rho (h^T_{i, k}, h^T_{i, k'})=\frac{Cov(h^T_{i, k}, h^T_{i, k'})}{\sqrt{D(h^T_{i, k})\cdot D(h^T_{i, k'})}}
	\end{align}
	where $Cov(\cdot)$ and $D(\cdot)$ represent the covariance and the variance between vectors respectively. Identically, the visual independent loss as the second part is computed as:
	\begin{equation}
		{\ell}^V_{ind}=\sum\nolimits_{i=1}^m \sum\nolimits_{k=1}^K \sum\nolimits_{k'=k+1}^K \rho (h^V_{i, k}, h^V_{i, k'})
	\end{equation}

The third part is cross-modal grounding loss. For a phrase $p_i$, the area of best-matched result $b_i$ in $\mathcal{B}$ is maximum,  overlapping with the ground truth $b_i^\text{\tiny +}$. We denote $\{\mathcal{B}-b_i\}$ as $M-1$ failed results. For outputs of the interventional model, we change their representations and treat them unmatched with ground truth. In total, there are $2M-1$ negative samples. Instead of minimizing the negative log-likelihood of correct correspondence scores, we consider a contrastive loss function, called InfoNCE \cite{oord2018representation}, in this paper:
	\begin{equation}
		\ell_{ground}=\sum\limits_{i=1}^n-\text{log}\frac{\text{exp}(\text{sim}(p_i, b_i)/\tau)}{\sum\nolimits_{j=1}^{2M-1}\text{exp}(\text{sim}(p_i, b_j)/\tau)}
	\end{equation}
	where $\text{sim}(\cdot)$ is the output of similarity prediction, the sum is over one positive and $2M-1$ negative regions, $\tau$ denotes the temperature parameter. During training process, for any phrase $p_i$, the model is tuned to maximize the numerator of the log argument and minimize its denominator as well. As a result, the model can thoroughly learn the difference between the true one and interventional samples.
	
Overall, all the losses are optimized to update learnable parameters:
	\begin{equation}
		\mathcal{L}=\ell^T_{ind}+\ell^V_{ind}+\ell_{ground}
	\end{equation}

\section{Experiments and Results}
\subsection{Datasets and Evaluation}
We validate our model on two common datasets for phrase grounding. 

Flickr30K Entities \cite{plummer2015flickr30k} contains 31,783 images where each image corresponds to five captions with annotated noun phrases. Consistent with the work \cite{dogan2019neural}, when a single phrase is annotated with multiple ground-truth bounding-box, we merge the boxes and use the union region as their ground-truth. We divide the dataset into 30k images for training, 1k for validation, and 1k for testing. 

ReferIt Game \cite{kazemzadeh2014referitgame} contains 20,000 images along with 99,535 segmented image regions. Each image is equipped with multiple referring phrases and corresponding bounding-boxes. We use the same split as \citet{akbari2019multi}, which contains 10k training and 10k test images. And we add the supplementary experiment on the Ref-COCO+ in Appendix D.

A noun phrase in the dataset is grounded correctly if and only if the predicted box and its ground-truth have at least 0.5 IoU (intersection over union). Based on the criteria, our measure of performance is grounding accuracy, which is the ratio of correctly grounded noun phrases to the total number of phrases in the test set.

\subsection{Experimental Setup}
We use the uncased Bert\footnote[1]{https://pypi.org/project/pytorch-pretrained-bert/} to get 768-dim phrase embeddings and a java toolkit\footnote[2]{https://nlp.stanford.edu/software/scenegraph-parser.shtml} to parse phrase scene graph for each caption. For visual scene graph generation\footnote[3]{https://github.com/rowanz/neural-motifs}, we use Faster R-CNN with VGG-16 backbone as a mechanism to extract top-100 proposal regions with 2048-dimension features and top-500 relations. All the dimensions of $d^T_{\text{in}}$, $d^V_{\text{in}}$, $d^T_{\text{out}}$ and $d^V_{\text{out}}$ are set as 512. For phrase and visual disentangled graph networks, the number of neighbor rooting layer is 2. We do a series of experiments to study the impact of the chunked number K in \{1, 2, 4, 8, 16\} and evaluate on two datasets. The results on Flickr30K is 77.44\%, 78.26\%, 78.73\%. 78.56\% and 78.08\% respectively. We can see our model reaches the best performance at 4. The accuracy on ReferIt is 63.94\%, 64.34\%, 65.15\%, 64.14\% and 64.83\% respectively. It is observed that the accuracy fluctuates after the k set to 4. Considering effectiveness and efficiency, we set K to 4. We only use one layer of Transformer with 4 multi-heads as cross-modal mapping. For the InfoNCE loss, the hyperparameter $\tau$ is set to 0.2. We train the end-to-end network by the SGD optimizer with learning rate 1e-3, weight decay 1e-4, and momentum 0.9. The model iterates 6 epochs with batch size 32.

\begin{table}[t]
	\small
	\renewcommand\arraystretch{1.2}
	\renewcommand\tabcolsep{1.0pt}
	\label{tab1}
%	\resizebox{\columnwidth}{!}{
	\begin{tabular}{cccc}
		\toprule[1pt]
		Model Type & Method & \textbf{Flickr30k} & \textbf{ReferIt}\\
		\midrule
		\multirow{14}{*}{\textbf{Independent}} & SMPL \cite{wang2016structured} & 42.08 & -\\
		& NonlinearSP \cite{wang2016learning} & 43.89 & -\\
		& SCRC \cite{hu2016natural} & - & 17.93\\
		& GroundeR \cite{rohrbach2016grounding} & 47.81 & 26.93\\
		& MCB \cite{fukui2016multimodal} & 48.69 & -\\
		& RtP \cite{plummer2015flickr30k} & 50.89 & -\\
		& SimNet \cite{wang2018learning} & 51.05 & 31.26\\
		& CGRE \cite{luo2017comprehension} & - & 31.85\\
		& MSRC \cite{chen2017msrc} & 57.53 & 32.21\\
		& IGOP \cite{yeh2017interpretable} & 53.97 & 34.70\\
		& SPC+PPC \cite{plummer2015flickr30k} & 55.49 & -\\
		& CITE \cite{plummer2018conditional} & 59.27 & 34.15\\
		& MultiG \cite{akbari2019multi} & 69.19 & 62.76\\
		& ZsgNet \cite{sadhu2019zero} & 63.39 & 58.63 \\
		& OneStage \cite{yang2019fast} & 68.69 & 59.30\\
		& SSNs \cite{plummer2020shapeshifter} & 72.50 & 60.50\\
		\midrule
		\multirow{2}{*}{\textbf{Sequential}} & QRC Net \cite{chen2017query} & 65.14 & 44.07\\
		& SeqGROUND \cite{dogan2019neural} & 61.60 & -\\
		\midrule
		\multirow{3}{*}{\textbf{Graph}} & GG++ \cite{bajaj2019g3raphground} & 66.93 & 44.91\\
		& VisualBERT \cite{li2020does} & 71.33 & -\\
		& LCMCG \cite{liu2020learning} & 76.74 & -\\
		& \textbf{DIGN} & \textbf{78.73} & \textbf{65.15}\\
		\bottomrule[1pt]
	\end{tabular}
%	}
	\renewcommand{\captionlabeldelim}{. }
	\caption{Overall accuracy(\%) comparison on two datasets}
	\vspace{-0.3cm}
\end{table}

\subsection{Results and Comparison}
The phrase grounding results of two benchmarks are reported in Table \ref{tab1}. We classify prior works and ours into three classes (i.e. independent, sequential and graph models) discussed in Section \ref{phrase_related}. Our model surpasses all state-of-the-art techniques and achieves the best performances of 78.73\% and 65.15\% on Flickr30K and ReferIt respectively. It can be apparently seen that our method achieves an absolute increase of $\sim$2\% compared to LCMCG on Flickr30K. Although both methods consider the context, our model learns the disentangled motif-aware context in scene graphs, which is the reason why our model can be improved. Comparing to MultiGrounding on ReferIt, we argue that the disentangled visual context and interventional sample strategies are mainly gain of improvement about $\sim$2.4\%. The fine categories comparison is analyzed in Appendix C.

\subsection{Ablation Studies}
\begin{table}[t]
	\small
	\centering
	\renewcommand\arraystretch{1.2}
	\renewcommand\tabcolsep{2.0pt}
	\label{tab4}
%	\resizebox{\columnwidth}{!}{
	\begin{tabular}{ccc}
		\toprule[1pt]
		Models & Flickr30K & ReferIt\\
		\midrule
		MLPs + MLPs Fusion & 65.07 & 35.46 \\
		GraphAttentionNetwork + MLPs Fusion & 72.44 & 60.31\\
		DisentangledGraphNetwork + MLPs Fusion & 74.34 & 61.71\\
		DGN + Cross-Modal Transformer & 75.15 & 62.74\\
		DGN + CMT + Structure Intervention  & 77.44 & 63.87\\
		DGN + CMT + Feature Intervention & 76.86 & 64.35\\
		\midrule
		DIGN & \textbf{78.73 }& \textbf{65.15}\\
		\bottomrule[1pt]
	\end{tabular}
%	}
	\renewcommand{\captionlabeldelim}{. }
	\caption{Ablation results on Flickr30k and ReferIt datasets.}
	\vspace{-0.3cm}
\end{table}
Extensive experiments are conducted to investigate the benefits of each block, with results enclosed in Table \ref{tab4}. To be more specific, the effectiveness of the disentangled graph network, two kinds of interventional strategies, and the cross-modal Transformer are clarified as follows. 

\begin{figure}[t]
	\centering
	\includegraphics[width=1\columnwidth]{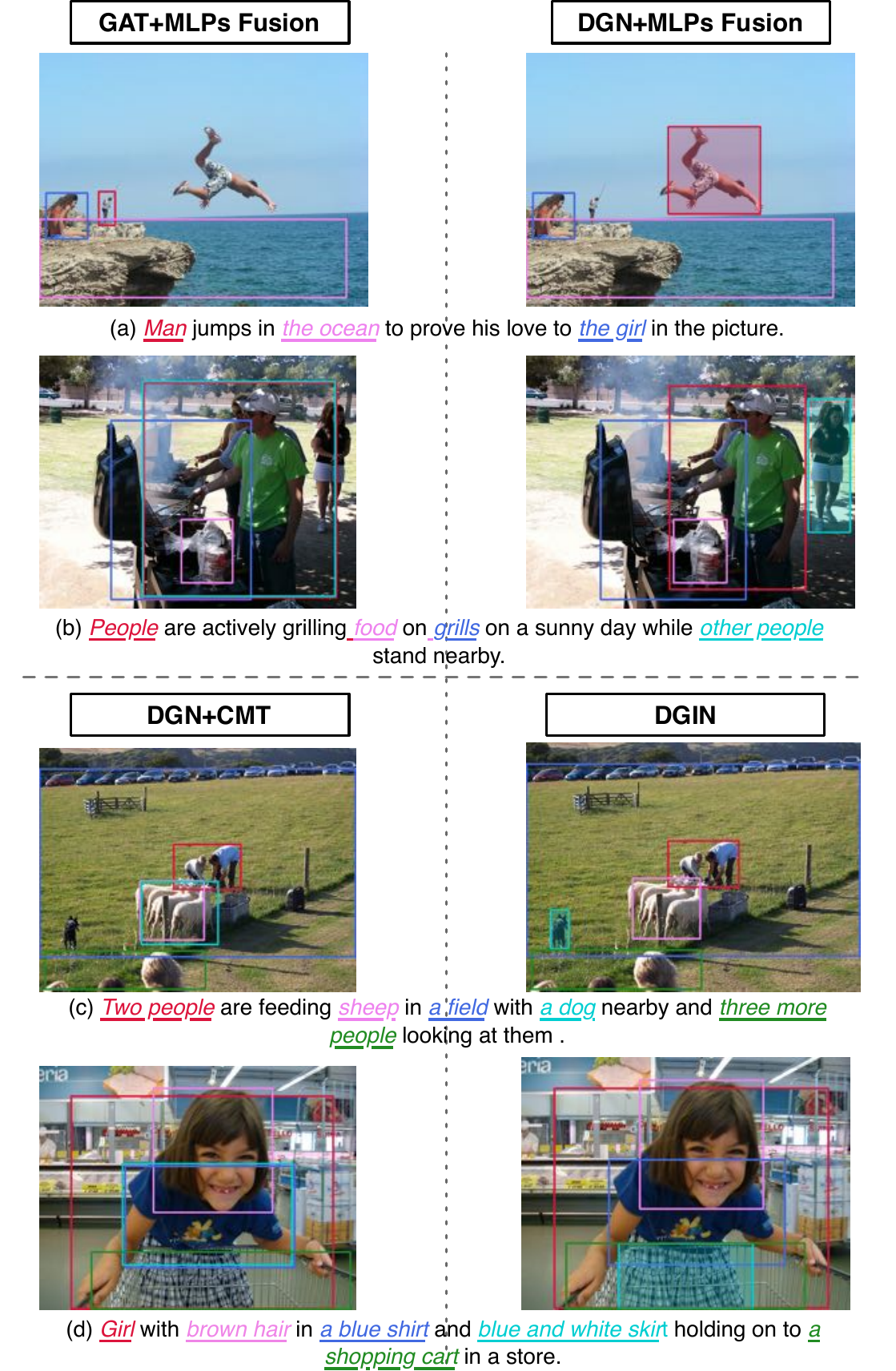}
	\renewcommand{\captionlabeldelim}{. }
	\vspace{-0.3cm}
	\caption{The visualization examples of models.}
	\vspace{-0.3cm}
	\label{fig_ablation}
\end{figure}
\begin{figure}[t]
	\centering
	\includegraphics[width=1\columnwidth]{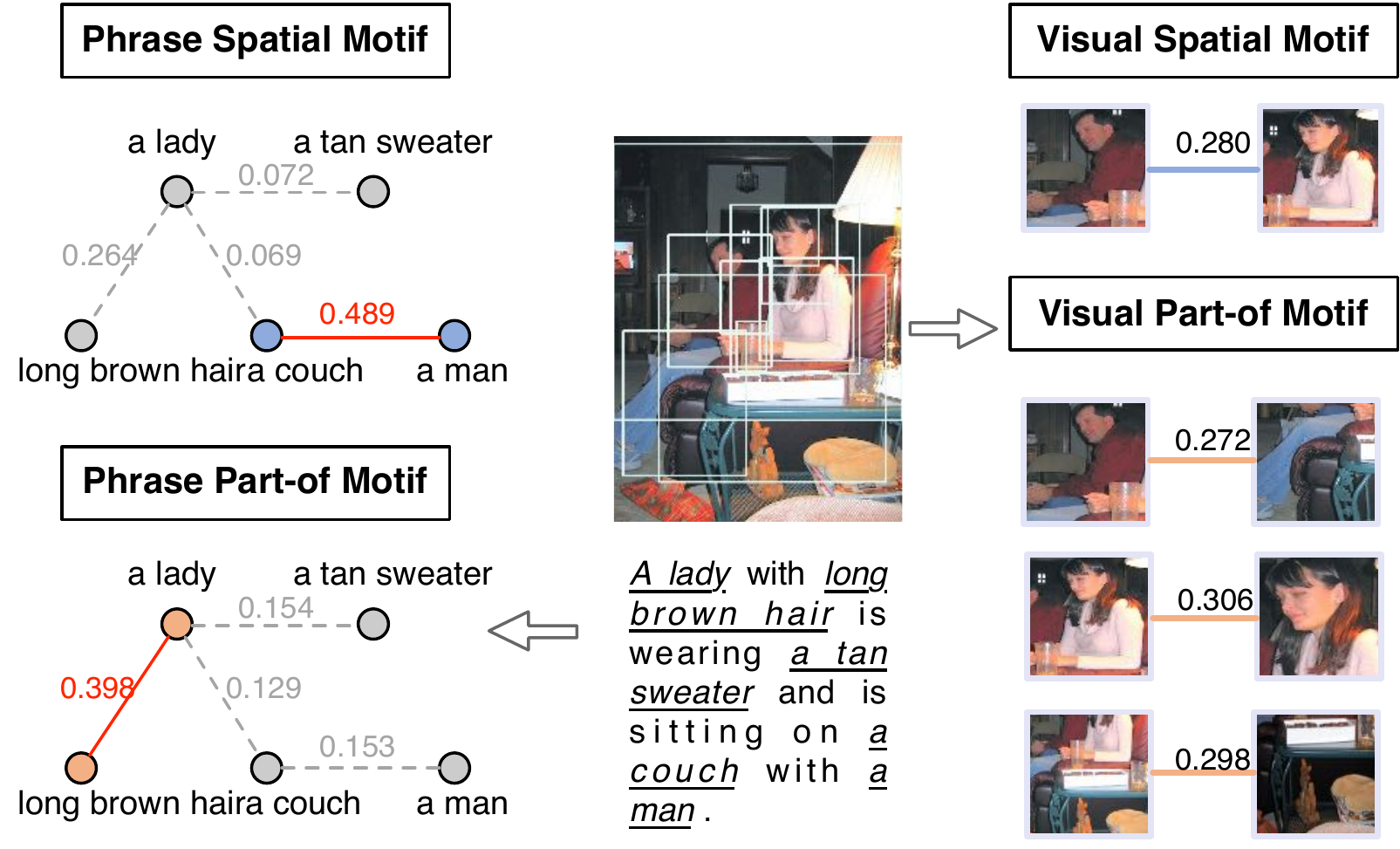}
	\renewcommand{\captionlabeldelim}{. }
	\vspace{-0.3cm}
	\caption{Illustration of motif-aware graph learning.}
	\vspace{-0.3cm}
	\label{fig:text_vis}
\end{figure}

The first line is used to validate the performance of our model is from our model design rather than better pre-trained  extractor. The baseline (the second line) utilizes the graph attention mechanism (GAT) in scene graphs and MLPs in inter-modal fusion for grounding. Disentangled graph network (DGN) is substituted for GAT to learn disentangled embeddings, which boosts the performance from 72.44\% to 74.34\% and from 60.31\% to 61.71\% on Flickr30K and ReferIt respectively. The third line replaces MLPs with a cross-modal Transformer (CMT) to transform $K$ independent components of intra-modal embedding to common subspaces and achieve alignment, improving accuracy within 0.8\%$\sim$1\%. The fourth and fifth lines are added to the third one by intervening in structure and feature levels respectively. Against the third line, the fourth one enhances the performance from 75.15\% to 77.44\% on Flickr30K and from 62.74\% to 63.87\% on ReferIt. In contrast to the third one, the fifth line has 1.7\% and 1.6\% impact on two benchmarks. Particularly, the performance of the structure intervention on Flickr30K is more influential than that of in the performance than the feature intervention. Because the number of images on Flickr30K is more than that on ReferIt, it is reasonable that the model on Flickr30K grasps more rich and complex context. The last line is our whole model and obtains the highest score across datasets. The main improvement can be shown from the second line to the second line (1.4\%$\sim$1.9\%) and from the third line to the last line (2.4\%$\sim$3.6\%). It is strongly proved that DGN and interventional methods are the most critical modules. In summary, The results show consistent patterns of effects in two datasets, showcasing that our model can generalize for the phrase grounding tasks. 

\subsection{Qualitative results}
From Figure \ref{fig_ablation}, we can see the contributions of the disentangled graph network module and interventional module. The upper part of the figure compares the graph attention network with the disentangled graph network, which proves the disentangled module can discriminate similar objects by chunked representations rather than messed representations. The lower part provides some interesting data, in terms of the fact that the interventional module can reduce the bias and break spurious relations at the structure level (e.g. the dog may usually co-occur with people and streets instead sheep) and recognize slight differences between objects at the feature level (e.g. the features of ``blue shirt" and ``blue and white skirt" are replaced with others to learn intrinsic color character).

We empirically inspect what kind of motifs are picked up by visualizing the weights of neighborhood routing. Figure \ref{fig:text_vis} shows the post-hoc explanations of learned spatial motif and part-of motif. There are more details in Appendix D. The left column shows two latent motifs of context in the phrase scene graph, and the right column exhibits motifs corresponding to phrase ones in the visual scene graph. For the phrase subgraph, the colored nodes represent that they are the most related in the current motif against other motifs. There are too many relationships in the visual scene graph, which are difficult to show a complete graph in the figure. So we just visualize the most prominent motif context between chunked nodes by the colored edges. From the results in Figure \ref{fig:text_vis}, the motif-aware context incorporated by the visual graph is more than that in the phrase graph even though disentangled. Hence, it is reasonable to compute the similarity of corresponding chunked embeddings to achieve grounding. Additional grounding results of our model are provided in Appendix E.

\section{Conclusion}
In this paper, we propose a disentangled interventional motif-aware learning framework for the phrase grounding task. Disentangled graph network distinguishes the importance of different motifs, integrates motif-aware context into node representations in intra-modal scene graphs, and provides interpretable results. Then interventional schemes at the aspects of feature and structure improve the resilience to the biased data and the ability of generalization. Finally, fine-grained intra-modal representations are fused in the cross-modal Transformer to finish phrase grounding. Our method (DGIN) is demonstrated on two public datasets, outperforming state-of-the-art by a considerable increase. In our future work, these tasks, constructing a more accurate scene graph and grounding phrase, should be considered to enhance each other. Additionally, we would like to make in-depth analyses of cross-modal fusing.

\section*{Acknowledgments}
This work has been supported in part by National Key Research and Development Program of China (2018AAA010010), Zhejiang NSF (LR21F020004), Funds from City Cloud Technology (China) Co. Ltd., Zhejiang University-Tongdun Technology Joint Laboratory of Artificial Intelligence, Zhejiang University iFLYTEK Joint Research Center, Chinese Knowledge Center of Engineering Science and Technology (CKCEST), Hikvision-Zhejiang University Joint Research Center, the Fundamental Research Funds for the Central Universities, and Engineering Research Center of Digital Library, Ministry of Education.

\bibliography{MuZ_DGIN.bib}

\begin{appendix}
\setcounter{secnumdepth}{1}
\section{Visual Disentangled Graph Network}
The visual disentangled graph network is clarified in the Algorithm \ref{vis_alg}. The visual embeddings firstly transform into $K$ subspaces (the 2nd-4th lines) to initialize the states of $K$ motif-aware subgraphs. We then build $L$ graph disentangling layers with the neighbor routing and updating mechanisms,  which integrate each kind of motif-aware context into related chunked embeddings (the 5th-17th lines).

\begin{algorithm}[h]
	\caption{Visual Disentangled Graph Network}
	\label{vis_alg}
	\begin{algorithmic}[1]
		\renewcommand{\algorithmicrequire}{\textbf{Input:}} 
		\REQUIRE $v_i\in\mathbb{R}^{d^V_{\text{in}}}$ and $j\in \mathcal{N}_i$
		\renewcommand{\algorithmicrequire}{\textbf{Output:}}
		\REQUIRE $h^V_i\in\mathbb{R}^{d^V_{\text{out}}}$		
		%\renewcommand{\algorithmicrequire}{\textbf{Params:}}
		%\REQUIRE ${W_k^V}^0\in \mathbb{R}^{\frac{d^V_{out}}{K} \times d^V_{in}}$, ${b^V_k}^0\in \mathbb{R}^{{\frac{d^V_{out}}{K}}}$, ${W^V_{ek}}^l$ and ${W^V_{k}}^l$ $\in \mathbb{R}^{\frac{d^V_{out}}{K} \times \frac{d^V_{out}}{k}}$.
		\FOR{$v_i \in \mathcal{V}^V$}
			%\STATE //project $t_i$ into the $k$-th subspace
			\FOR{chunk $k=1,...,K$}
				\STATE $v^0_{i, k} \gets \sigma({W^V_k}^0 v_i+{b^V_k}^0)/\left\| \sigma({W^V_k}^0 v_i+{b^V_k}^0) \right\|_2$ 
			\ENDFOR
			\FOR{ graph disentangling layer $l=1,...,L$}
				%\STATE // neighbor rooting
				\FOR{$v_j$ that satisfies $j \in \mathcal{N}_i$}
					\STATE ${a^V_{j, k}}^l \gets {v^{l-1}_{j, k}}^\intercal v^{l-1}_{i, k}$, $\forall k=1,...,K$
					\STATE $[{a^V_{j, 1}}^l,...,{a^V_{j, K}}^l] \gets softmax([{a^V_{j, 1}}^l,...,{a^V_{j, K}}^l])$ 
				\ENDFOR
				\FOR{component $k=1,...,K$}
					%\STATE // message passing
					\STATE 
						${A^V_{i,k}}^l \gets \sum\nolimits_{j\in \mathcal{N}_i}{a^V_{j, k}}^l {W^V_k}^l v^{l-1}_{j, k}$
					%\STATE // node update
					\STATE $v^{l}_{i,k} \gets \sigma({W^V_{ek}}^{l}v^{l-1}_{i, k}+{A^V_{i,k}}^l)$
				\ENDFOR
				\STATE $v^{l}_i \gets [v^{l}_{i,1},...,v^{l}_{i,K}]$
			\ENDFOR
			%// the final output of the phrase graph node $i$
			\STATE $h^V_i \gets v^{0}_i+...+v^{L}_i$
		\ENDFOR
	\end{algorithmic}
\end{algorithm}

\section{Interventional Visual representation}
Algorithm \ref{alg2} shows the overall pseudocode of the interventional algorithm at the feature and the structure levels in the visual disentangled graph network (VDGN). For each training sample, we randomly use one specific synthesizing mechanism, and $\delta$ is the trade-off weight.

\begin{algorithm}[h]
	\begin{algorithmic}[1]
		\renewcommand{\algorithmicrequire}{\textbf{Input:}} 
		\REQUIRE nodes $\mathcal{V}^T$ and edges $\mathcal{E}^T$ of visual scene graph.
		\renewcommand{\algorithmicrequire}{\textbf{Output:}} 
		\REQUIRE the interventional features of nodes.
		\STATE $cond \sim U[0,1]$
		\IF{$cond \geq \delta$}
			%\STATE // relation intervene
			\STATE $\{v^0_1,...,v^0_m\} \gets \text{map}\ \mathcal{V}^V \text{into}\ K\ \text{subspaces}$
			\STATE $H^V_{neg} \gets \text{VDGN}(\{v^0_1,...,v^0_m\},\mathcal{E}^V_{neg})$
		\ELSE
			%\STATE // feature intervene
			\STATE $\{v^0_1,...,v^0_m\} \gets \text{map}\ \mathcal{V}^V \text{into}\ K\ \text{subspaces}$
			\STATE $\text{random choice k} \in [1,K]$
			\FORALL {$v^0_i \in \{v^0_1,...,v^0_m\}$}
			\STATE $v^0_{i_{neg}} \gets \{v^0_{i,1},...,v^0_{i,k_{neg}},...,v^0_{i,K}\}$
			\ENDFOR
			\STATE $H^V_{neg} \gets \text{VDGN}(\{v^0_{1_{neg}},...,v^0_{m_{neg}}\},\mathcal{E}^V)$
		\ENDIF
	\end{algorithmic}
	\caption{Interventional Visual Sample Synthesizing} 
	\label{alg2}
\end{algorithm}

\section{Category-wise accuracy comparesion}
For a more in-depth understanding, Table \ref{tab2} shows the models per category's performance on Flickr30K. Learning discriminative motif-ware context contributes to the success of some categories, such as ``people", ``clothing", ``body parts", ``scene" and ``other".  It is reasonable because ``people" does share some motifs like decorative, part-of, action with the other categories mentioned above. For the sparse data of ``animals", ``vehicles" and ``instruments" categories, most of them increase steadily owing to interventional schemes. When intervening in the model, we provide more unseen environments and make representations robust and general.
\begin{table}[h]
	\caption{Category-wise accuracy on Flickr30K test dataset}
	\renewcommand\arraystretch{1.2}
	\label{tab2}
	\resizebox{\columnwidth}{!}{
	\begin{tabular}{ccccccccc}
		\toprule[1pt]
		Method & people & clothing & body parts & animals & vehicles & instruments & scene & other\\
		\midrule
		SMPL & 57.89 & 34.61 & 15.87 & 55.98 & 52.25 & 23.46 & 34.22 & 26.23\\
		GroundeR & 61.00 & 38.12 & 10.33 & 62.55 & 68.75 & 36.42 & 58.18 & 29.08\\
		RtP & 64.73 & 46.88 & 17.21 & 65.83 & 68.72 & 37.65 & 51.39 & 31.77\\
		IGOP & 68.71 & 56.83 & 19.50 & 70.07 & 73.75 & 39.50 & 60.38 & 32.45\\
		SPC+PPC & 71.69 & 50.95 & 25.24 & 76.23 & 66.50 & 35.80 & 51.51 & 35.98\\
		CITE & 73.20 & 52.34 & 30.59 & 76.25 & 75.75 & 48.15 & 55.64 & 42.83\\
		SeqGROUND & 76.02 & 56.94 & 26.18 & 75.56 & 66.00 & 39.36 & 68.69 & 40.60\\
		QRC Net & 76.32 & 59.58 & 25.24 & 80.50 & 78.25 & 50.62 & 67.12 & 43.60\\
		GG++ & 78.86 & 68.34 & 39.80 & 81.38 & 76.58 & 42.35 & 68.82 & 45.08\\
		MultiGrounding & 75.60 & 58.30 & 44.90 & 87.60 & 83.80 & 57.50 & 68.20 & 59.80\\
		LCMCG & 86.82 & \textbf{79.92} & 53.54 & \textbf{90.73} & 84.75 & 63.58 & 77.12 & 58.65\\
		\midrule
		\textbf{DIGN} &\textbf{86.95} & 78.09 &\textbf{61.46} & 86.54 &\textbf{86.63} &\textbf{65.73} &\textbf{81.94} &\textbf{66.71}\\
		\bottomrule[1pt]
	\end{tabular}}
\end{table}

\section{The Ref-COCO+ Result}
	Our model is tested on Ref-COCO+ and achieves 70.21\% accuracy, which surpasses MattNet (64.93\%) and CM-Att-Erase\footnote[1]{Xihui Liu, et al. Improving referring expression grounding with cross-modal attention-guided erasing. In CVPR, 2019.} (68.09\%). However, there is a slight disparity in comparison with VLBert\footnote[2]{Su, Weijie, et al. Vl-bert: Pre-training of generic visual-linguistic representations. In ICLR, 2019.} (71.84\%), ViLBert\footnote[3]{Lu, J., et al. Vilbert: Pretraining task-agnostic visiolinguistic representations for vision-and-language tasks. In NeurIPS, 2019.} (72.34\%), UNITER\footnote[4]{Chen, Y. C., et al. Uniter: Learning universal image-text representations. In arXiv:1909.11740.} (74.94\%). The cross-modal Transformer series pretrain on large scale V+L datasets like Concept Caption to learn more prior knowledge and get better results.

\section{Motif-aware Visualization Results}
\begin{figure*}[!t]
	\centering
	\includegraphics[width=1\textwidth]{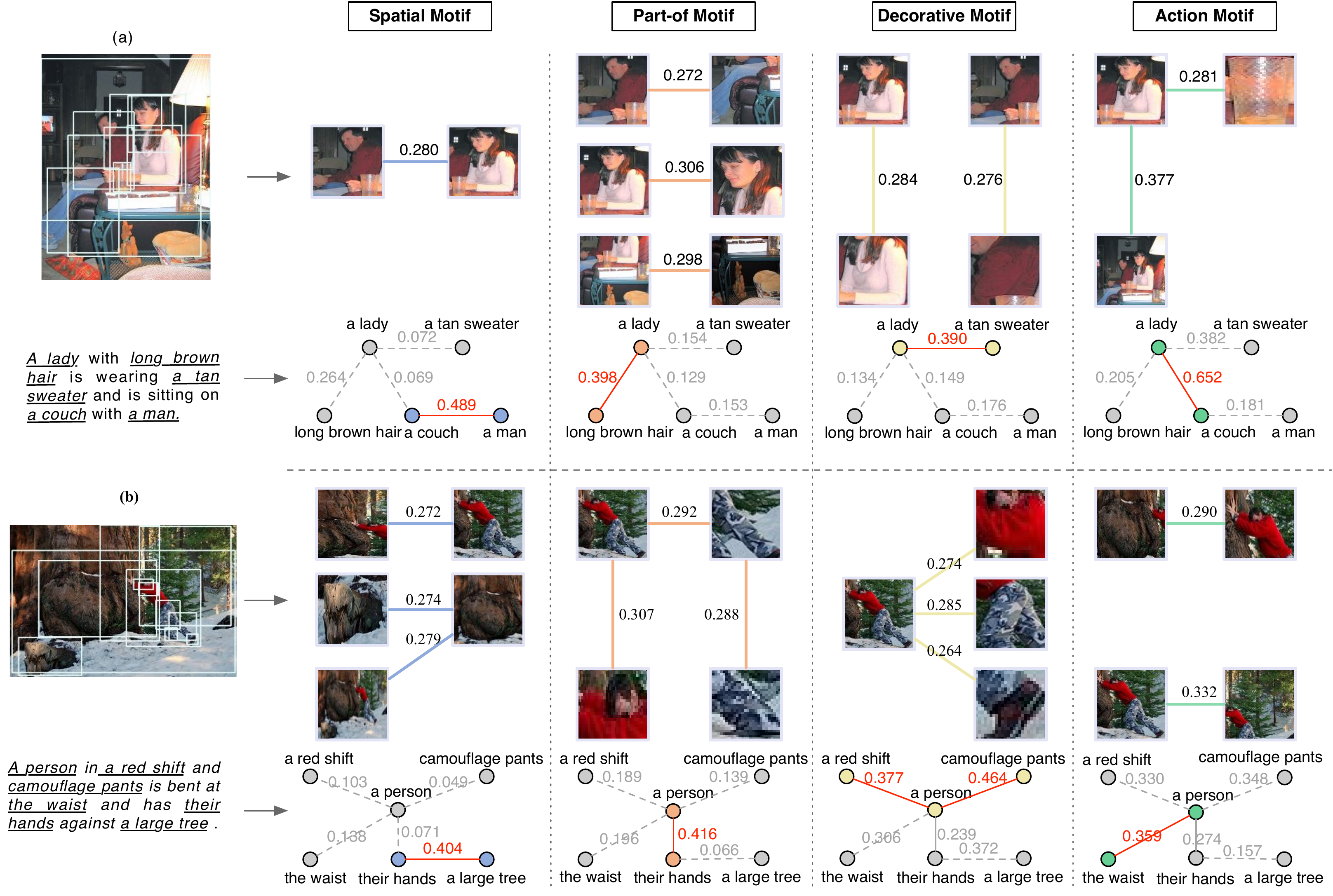}
	\caption{Illustration of motif-ware graph disentangling layer in both scene graphs. Edges' colors represent different motifs in the visual graph. Different nodes' color show chunked embeddings learned different motifs in the phrase graph.}
	\label{fig:all_disen}
\end{figure*}

Intact learned motif examples are depicted in Figure \ref{fig:all_disen} in phrase and visual scene graphs respectively. We set the hyper-parameter of latent motifs as 4, and get four kinds of post-hoc interpretable motifs in the context of the graph.

\begin{figure*}[!t]
	\centering
	\includegraphics[width=1\textwidth]{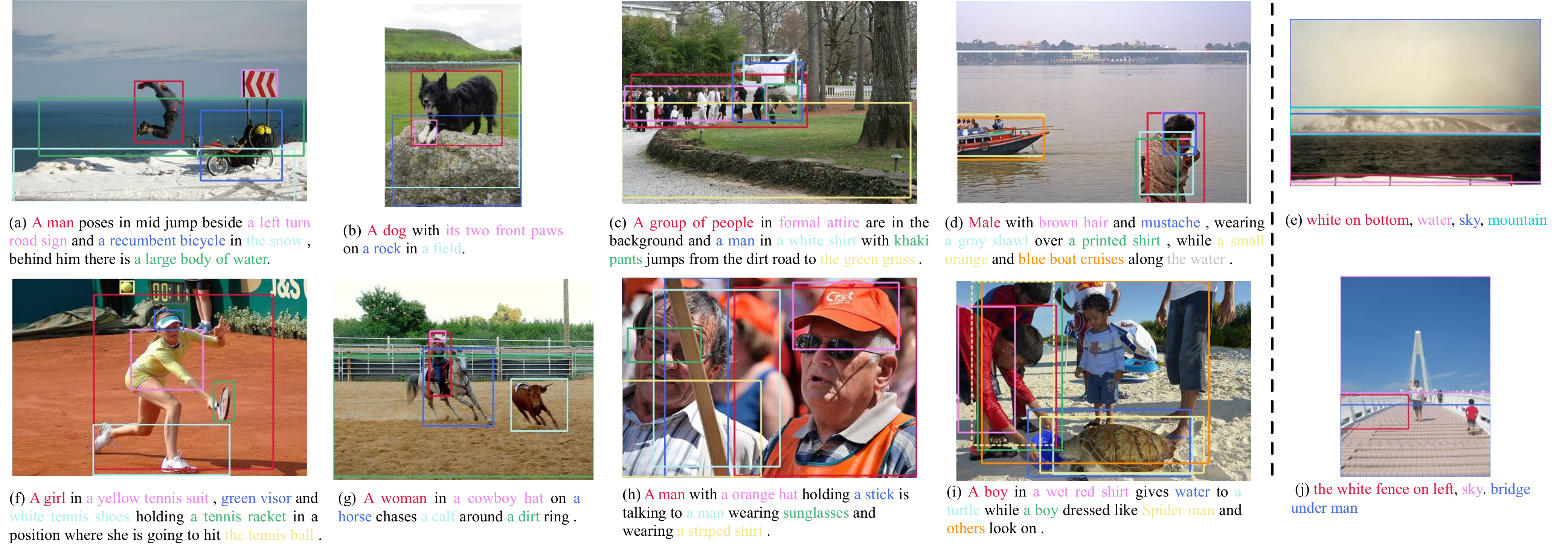}
	\caption{Qualitative results on Flickr30K (the first four columns) and ReferIt (the last column). The colored phrases of captions are grounded to regions in same color.}
	\label{fig:flickr_vis}
\end{figure*}

\section{Grounding Results}
In Figure \ref{fig:flickr_vis}, the results of phrase grounding  demonstrate the following abilities. Firstly, our model is able to ground multiple phrase to corresponding regions such as (a), (b), (f), (g). Second, the situation of two ambiguous instances in the sentence is solved by our disentangled model (e.g., the white shirt in (c) and the sunglasses in (h) exist some matching bounding boxes). The blue dotted ground truth in (d) and the dotted yellow box in (i) are failed to be grounded. On the one hand, it is not easy to recognize even humans for such a mustache's tinny size. On the other hand, common sense like the spiderman wearing red striped cloth is not involved in our model.
\end{appendix}
\end{document}